\documentclass[journal]{IEEEtran}

\usepackage{cite}
\usepackage{color}

\usepackage[dvips]{graphicx}

\ifCLASSINFOpdf
\else
\fi
\usepackage[cmex10]{amsmath}
\usepackage{algorithm}
\usepackage{algorithmic}

\usepackage{color}

\usepackage{multicol}
\usepackage{multirow}

\usepackage[caption=false,font=footnotesize]{subfig}

\begin{document}
%
\title{Convolution in Convolution for Network in Network}
%
%
%

\author{Yanwei~Pang,~\IEEEmembership{Senior~Member,~IEEE,}
        Manli~Sun,
        Xiaoheng~Jiang,
        and~Xuelong~Li,~\IEEEmembership{Fellow,~IEEE}
\thanks{Y. Pang, M. Sun, and X. Jiang are with the School of Electronic Information Enginnering, Tianjin University, Tianjin 300072, P. R. China. (E-mail: pyw@tju.edu.cn, sml@tju.edu.cn, jiangxiaoheng@tju.edu.cn).}
\thanks{X. Li is with the Center for OPTical IMagery Analysis and Learning (OPTIMAL), State Key Laboratory of Transient Optics and Photonics, Xi'an Institute of Optics and Precision Mechanics, Chinese Academy of Sciences, Xi'an 710119, Shaanxi, P. R. China. E-mail: xuelong{\_}li@opt.ac.cn}
}

\maketitle

\begin{abstract}
Network in Netwrok (NiN) is an effective instance and an important extension of Convolutional Neural Network (CNN) consisting of alternating convolutional layers and pooling layers. Instead of using a linear filter for convolution, NiN utilizes shallow MultiLayer Perceptron (MLP), a nonlinear function, to replace the linear filter. Because of the powerfulness of MLP and $ 1\times 1 $ convolutions in spatial domain, NiN has stronger ability of feature representation and hence results in better recognition rate. However, MLP itself consists of fully connected layers which give rise to a large number of parameters. In this paper, we propose to replace dense shallow MLP with sparse shallow MLP. One or more layers of the sparse shallow MLP are sparely connected in the channel dimension or channel-spatial domain. The proposed method is implemented by applying unshared convolution across the channel dimension and applying shared convolution across the spatial dimension in some computational layers. The proposed method is called CiC. Experimental results on the CIFAR10 dataset, augmented CIFAR10 dataset, and CIFAR100 dataset demonstrate the effectiveness of the proposed CiC method. 
\end{abstract}


%
\IEEEpeerreviewmaketitle

\section{Introduction}
%
%
%
%
\IEEEPARstart{D}{EEP} Convolutional Neural Networks (CNNs) have achieved state-of-the-art performance in the tasks of image recognition and object detection. CNNs are organized in successive computational layers alternating between convolution and pooling (sub-sampling). Compared to other types of deep neural networks, CNNs are relatively easy to train with back-propagation mainly because they have a very sparse connectivity in each layer \cite{Bengio_Learningdeeparchitectures_FTML2006}. In a convolutional layer, linear filters are used for convolution. The main parameters of CNNs are the parameters (i.e., weights) of the filters. To reduce the number of parameters, a parameter sharing strategy is adopted. Although parameter sharing reduces the capacity of the networks, it improves its generalization ability (Sec. 4.19, \cite{Haykin_NeuralNetworks_1999}). The computational layers can be enhanced by replacing the linear filter with a non-linear function: shallow MultiLayer Perceptron (MLP) \cite{Lin_NiN_CoRR2013}. The CNN with shallow MLP is called NiN \cite{Lin_NiN_CoRR2013}. With enough hidden units, MLP can represent arbitrary complex but smooth functions and hence can improve the separability of the extracted features. So NiN is able to give lower recognition error than classical CNN. 

As a filter in CNNs, a shallow MLP convolves across the input channels. Because the filter itself is also a network, the resulting CNN is called Network in Network (NiN). But the MLP in NiN does not employ sparse connectivity. Instead, MLP is a fully connected network. So many parameters of MLP have to be computed and stored. This limits the performance of NiN. To breakthrough the limitation, in this paper we propose to modify the fully connected MLP to a locally connected one. This is accomplished by leveraging a kernel (a.k.a., filter) on each layer (or some layers) of the MLP. That is, the size of the kernel is smaller than that of the input. Because the convolution operation is conducted in the embedded MLP of the convolution neural networks, we call the proposed method Convolution in Convolution (CiC). In summary, the contributions of the paper and the merits of the proposed CiC are as follows.
\begin{enumerate}

\item A fully sparse (locally connected) shallow MLP and several partial sparse shallow MLPs (e.g., MLP-010) are proposed and are used for convolutional filters. The convolutional filter itself is obtained by convolving a linear filter. 

\item We develop a CNN method (called CiC) with the sparse MLPs. In CiC, shared convolution is conducted in the spatial domain and unshared convolution is conducted along the channel dimension. 

\item The basic version (i.e., CiC-1D) of CiC utilizes $ 1\times 1 $ convolutions in spatial domain and applies one-dimensional filtering along the channel dimension. We then generalize CiC-1D into CiC-3D by replacing the $ 1\times 1 $ convolutions with $ n\times n $ convolutions. 

\item The proposed CiC method significantly outperforms NiN in reducing the test error rate at least on the CIFAR10 and CIFAR100 datasets.
\end{enumerate}

The rest of the paper is organized as follows. We review related work in Section II. The proposed method is presented in Section III. Subsequently, experimental results are provided in Section IV. We then conclude in Section V based on these experimental results.

\section{Related Work}
In this section, we will first briefly the basic components of the CNNs. Next, we will review some directions of CNNs which are related to our work. 

Generally, CNNs are mainly organized in interweaved layers of two types: convolutional layers and pooling (subsampling) layers \cite{Bengio_Learningdeeparchitectures_FTML2006,LeCun_Gradientbased_IEEE1998,Krizhevsky_Imagenetclassification_NIPS2012,Gong_AMultiobjective_TNN2015,Chang_Deepandshallow_TNN2015} with a convolutional layer (or several convolutional layers) followed by a pooling layer. The role of the convolutional layers is feature representation with the semantic level of the features increasing with the depth of the layers. Each convolutional layer consists of several feature maps (a.k.a., channels). Each feature map is obtained by sliding (convoluting) a filter over the input channels with predefined stride followed by a nonlinear activation. Different feature maps correspond to different parameters of filters with a feature map sharing the same parameters. The filters are learned with back propagation. Pooling is a process that replaces the output of its corresponding convolutional layers at certain location with summary statistic of the nearby outputs \cite{Bengio_Learningdeeparchitectures_FTML2006}. Pooling over spatial regions contribute to make feature representation become translation invariant and also contribute to improve the computational efficiency of the network. The layers after the last pooling layer are usually fully connected and are aimed at classification. The number of layers is called the depth of the network and the number of units of each layer is called the width of the network. The number of feature maps in each layer can also represent the width (breadth) of CNNs. The depth and width determines the capacity of CNNs.

Generally speaking, there are six directions to improve the performance of the CNNs with some of the them are overlapping: (1) increasing the depth; (2) increasing the width; (3) modifying the convolution operation \cite{Ciresan_Multicolumndeep_CoRR2012,Jarrett_Whatis_ICCV2009,LeCun_Convolutionalnetworks_ISCS2010,Krizhevsky_Imagenetclassification_NIPS2012,Gregor_Emergenceof_CoRR2010}; (4) modifying the pooling operation \cite{Zeiler_Stochasticpooling_CoRR2013,He_Spatialpyramid_ECCV2014,Chan_PCANet_CoRR2014,Lee_Generalizingpooling_CoRR2015,Springenberg_Strivingfor_CoRR2014,Gong_Multiscaleorderless_ECCV2014,Yoo_Multiscalepyramid_CVPR2015,Graham_FractionalMax_CoRR2014,Murray_Generalizedmax_ICML2010}; (5) reducing the number of parameters; and (6) modifying the activation function. Our method is closely related to NiN and NiN is relevant to the first three directions. 

(1) \textbf{Increase the depth.} Large depth of deep CNNs is one of the main differences between CNNs and traditional neural networks \cite{Shao_Learningdeepand_TNN2014,Chang_Deepandshallow_TNN2015}. LeNet-5 \cite{LeCun_Gradientbased_IEEE1998} is a seven-layer version of CNNs which has three convolutional layers, two pooling (subsampling) layers, and two fully connected layers. AlexNet \cite{Krizhevsky_Imagenetclassification_NIPS2012} contains eight learned layers (not accounting the pooling layers or taking a convolutional layer followed by a pooling layer as a whole) with five convolutional ones and three fully-connected ones. In VggNet \cite{Simonyan_Verydeep_CoRR2014}, the depth is up to 19. GoogLeNet \cite{Szegedy_Goingdeeper_CoRR2014} is a 22 layer CNNs. By using gating units to regulate the flow of information through a network, Highway Networks \cite{Srivastava_Highwaynetworks_CoRR2015} opens up the possibility of effectively and efficiently training hundreds of layers with stochastic gradient descent. In ResNet \cite{He_Deepresidual_CoRR2015}, the depth of 152 results in state-of-the-art performance. Zeiler \textit{et al.} showed that having a minimum depth to the network, rather than any individual section, is vital to the model’s performance \cite{Matthew_Visualizingand_ECCV2014}. By incorporating micro networks, NiN \cite{Lin_NiN_CoRR2013} also increases the depth. The depth of our method is same as that of NiN. 

(2) \textbf{ Increase the width.} Increasing the number of feature maps in a convolutional layer yields enriched features and hence is expected to improve the CNNs \cite{LeCun_Gradientbased_IEEE1998}. Zeiler \textit{et al.} found that increasing the size of the middle convolution layers gives a useful gain in performance \cite{Matthew_Visualizingand_ECCV2014}. GoogLeNet \cite{Szegedy_Goingdeeper_CoRR2014} is famous for not only its large depth but also its large width. In GooLeNet, a group of convolution filters of different sizes form an Inception module. Such an Inception module greatly increases the network width. The OverFeat network \cite{Sermanet_Overfeat_CoRR2013} utilizes more than 1000 feature maps in both the 
$4^{th}$ and $5^{th}$ convolutional layers. It is noted that large width implies large computation cost. Without increasing the width, our method outperforms NiN in terms of error rate. 

(3) \textbf{Modifying the convolution operation.} There are several ways to modify the convolution operation: changing the sliding stride, filter size, and filter type. It was found that a large convolution stride leads to aliasing artifacts \cite{Matthew_Visualizingand_ECCV2014}. Therefore, it is desirable to use small stride. Moreover, decreasing the filter size of the first convolution layer from $ 11\times 11 $ to $ 7\times 7 $ is positive for performance improvement. Rather than learning a set of separate set of weights at every spatial location, titled convolution \cite{Gregor_Emergenceof_CoRR2010} allows one to learn a set of filters that one rotate through as we move through space, \cite{Goodfellow_DeepLearning_2015}. Modifying the filter type is an important attempt to develop effective CNNs \cite{Matthew_Visualizingand_ECCV2014}. While most of methods employ linear filter, NiN \cite{Lin_NiN_CoRR2013} adopts a nonlinear filter: shallow Multiple Layer Perception (MLP) which significantly enhances the representational power of CNNs \cite{Szegedy_Goingdeeper_CoRR2014}. CSNet \cite{Jiang_CascadedSubpatch_CoRR2016} utilizes cascaded subpatch filters for convolution computation. Our method directly modifies NiN in the sense of convolutional layer.

\section{Proposed Method: CiC}
In this section, we present an improved NiN (Network in Network) \cite{Lin_NiN_CoRR2013} which we call CiC (Convolution in Convolution). One of the characteristics of CiC is that sparse shallow MLPs are used for convolutionally computing the convolutional layers and the shallow MLPs themselves are obtained by convolution. The proposed CiC is equivalent to apply unshared convolution across the channel dimension and applying shared convolution across the spatial dimension in some computational layers. We first describe the basic idea of CiC (i.e., CiC-1D) with sparse and shallow MLP and then extend it to three-dimensional case (i.e., CiC-3D).

\subsection{From dense shallow MLP to sparse shallow MLP}
In classical CNNs, linear filters are used for calculating the convolutional layers. However, there is evidence that more complex and nonlinear filter such as shallow MLP is preferable to the simple linear one \cite{Lin_NiN_CoRR2013}. In NiN \cite{Lin_NiN_CoRR2013}, dense (i.e., fully connected) MLP is used as a filter (kernel). In our method, we propose to modify the dense MLP (see Fig. 1(a) for an example) to sparse one (see Figs. 1(b)-(f) for examples). We divide sparse MLP into full sparse MLP and partial MLP. 

Fig. 1(a) is two-hidden-layer dense MLP which has 48, 24, and 8 free parameters (weights) in hidden layer 1 (the first hidden layer), hidden layer 2 (the second hidden layer), and output layer, respectively. Totally, there are 48+24+8=80 parameters in the dense MLP. Fig. 1(b) is a two-hidden-layer full sparse MLP which is obtained by convolving a linear filter with three weights (called inner filter) across each layer. The sparse MLP with parameter sharing mechanism has only three free parameters in each layer and 3+3+3=9 parameters in total. If unshared convolution is employed, then the number of parameters became 36. Therefore, the sparse MLP has fewer parameters than the dense counterpart. Few parameters is capable of reducing memory consumption, increasing statistical efficiency, and also reducing the amount of computation needed to perform forward and back-propagation \cite{Goodfellow_DeepLearning_2015}. 

In addition to full sparse MLP, partial sparse MLP can be used. Fig. 1(c)$\sim$(f) show several possible partial sparse MLPs. To distinguish the different partial spare MLPs, we use `1' to mean that a layer is locally connected and use `0' to mean that the layer is fully connected. A sequence of the labels is used for expressing a certain type of partial sparse MLP. For example, in Fig. 1(c) the hidden layer 1 is densely connected (labeled by `0'), hidden layer 2 is sparsely connected (labeled by `1'), and the output layer is densely connected (labeled by `0'). Therefore, we utilize ``010" to represent the specific type of the partial sparse MLP. Specifically, the MLP in Fig. 1(c) is called MLP-010. Similarly, the MLPs in Fig. 1(d), (e), and (f) are called, MLP-011, MLP-100, and MLP-110, respectively. 

\begin{figure}[!t]
\centering
\includegraphics[scale=1]{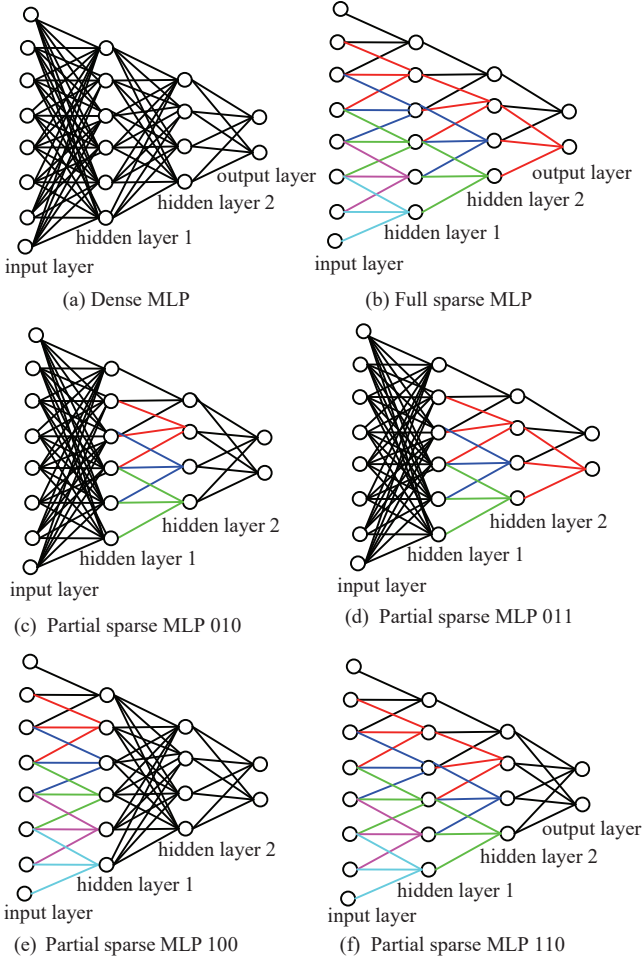}
\caption{Shallow Multi-layer Perception (MLP). (a) Dense (fully connected) MLP. (b) Full sparse MLP. (c)$\sim$(e) Different types of partial sparse MLPs.}
\label{Fig1}
\end{figure}

Our experimental results show that unshared convolution is better than shared convolution in the sense of reducing the test error rate. Therefore, unshared convolution is adopted along the channel dimension. Unshared convolution means that a set of different weights are used for sparse connection.

\subsection{CiC with Sparse and Shallow MLP (One-dimensional Filtering across Channels with Large Filter): CiC-1D}
As stated above, we propose to employ unshared convolution for computing sparse and shallow MLP. In our CiC method, the sparse MLP is regarded as a convolution filter and inserted into the framework of CNNs. 

Fig. 2(a) shows the proposed CiC with MLP-010. Assume that all the input color images are of size $W \times H$. As NiN, CiC also has three building blocks (i.e., block 1, block 2, and block 3). In block $i$, the input of an MLP-010 is of size $w_i  \times h_i$ in spatial domain and $L_0^i$ in channel domain. The numbers of the neurons in the first hidden layer, the second hidden lay, and the output layer of an MLP-010 in block $i$ are denoted by $N_1^i$, $N_2^i$, and $N_3^i$, respectively. 

Fig. 2(a) explicitly shows the role of MLP-010 in constructing CiC whereas Fig. 2(b) and (c) show the architecture of CiC in the manner of convolutional layers. Both Fig. 2(b) and (c) are equivalent to Fig. 2(a). Note that the pooling layers exist but are not shown in Fig. 2(b). In Fig. 2(b), the small cube (i.e., the dashed cube) inside a large cube (i.e., solid cube) stands for a filter (kernel). The kernel is a three-order tensor ${\bf{k}} \in R^{x \times y \times z}$ where $x$ and $y$ index the spatial domain (i.e., $x$ and $y$ are the length of the kernel in horizontal and vertical axis, respectively) and $z$ indexes the channel domain (i.e., $z$ is the length of the kernel in the channel axis). In each block, there are three kernels, corresponding to the input layer, the first hidden layer, and the second layer of an MLP-010. The kernels in block 1, block 2, and block 3 are ( ${{\bf{k}}_0}$, ${{\bf{k}}_1}$, ${{\bf{k}}_2}$), ( ${{\bf{k}}_3}$, ${{\bf{k}}_4}$, ${{\bf{k}}_5}$), and ( ${{\bf{k}}_6}$, ${{\bf{k}}_7}$, ${{\bf{k}}_8}$) (see Fig. 2(b)), respectively. If the number of output channels obtained by a kernel ${{\bf{k}}_i}$ is taken into account, the kernel can be expressed as a four-order tensor ${\bf{K}}_i  \in R^{x \times y \times z \times c}$ (see Fig. 2(c)) where the first three indices (i.e., $x$, $y$, and $z$) have the same meaning as that of ${{\bf{k}}_i}$ and the last index $c$ stands for the number of feature maps (channels) obtained by the kernel ${{\bf{k}}_i}$.

Now we dicuss the three kernels in block 1.

\textbf{The first kernel} ${{\bf{k}}_0}$(${\bf{K}}_0$): In the input layer of block 1 (also the input of the first MLP010), the kernel ${\bf{k}}_0  \in R^{w_1  \times h_1  \times L_0^1 }$ (or equivalently ${\bf{K}}_0  \in R^{w_1  \times h_1  \times L_0^1  \times N_1^1 }$) is used for filtering. The small numbers ${w_1}$ and ${h_1}$ are larger than 1 (e.g., ${w_1}$=5, ${h_1}$=5). Because the input is usually a color image, so the channel number ${L_0^1}$ equals to 3 (i.e., three color channels). Because the number of channels is very small, the channels have to be densely (fully) connected by using the kernel ${{\bf{k}}_0}$ with its size being ${w_1 \times h_1 \times 3}$. Denote ${N_1^1}$ the number of channels output by ${\bf{K}}_0$.

\textbf{The second kernel} ${{\bf{k}}_1}$(${\bf{K}}_1$): The output of the first kernel is the input of the second kernel ${{\bf{k}}_1}$(or ${\bf{K}}_1$) which is of size ${1 \times 1 \times L_1^1}$(or ${1 \times 1 \times L_1^1 \times N_2^1}$) with ${L_1^1 < N_1^1}$. The convolution in spatial domain with the ${1 \times 1}$ kernel plays an important role in dimensionality reduction \cite{Szegedy_Goingdeeper_CoRR2014,Lin_NiN_CoRR2013}. Because of ${L_1^1 < N_1^1}$(i.e., the length ${L_1^1}$ of the filter in channel dimension is smaller than the number ${N_1^1}$ of the input channels), the kernel ${{\bf{k}}_1}$ connects the ${N_1^1}$ channels in a convolutional (i.e., sparse) manner. In Section IV.B, we state how to choose the optimal kernel length in channel dimension. 

${N_2^1}$ is the number of channels output by applying the kernels ${\bf{K}}_1 \in R^{1  \times 1  \times L_1^1  \times N_2^1 }$. Usually, zero-padding is used for covolutional implementation \cite{Goodfellow_DeepLearning_2015}. However, throughout this paper, no padding is applied along the channel direction. Consequently, the number  ${N_2^1}$ of channels output by applying the kernels ${\bf{K}}_1 \in R^{1  \times 1  \times L_1^1  \times N_2^1 }$ is completely determined by the kernel length ${L_1^1}$ and the number ${N_1^1}$ of input channels. Specifically, we have:
\begin{equation}
\label{eq1}
N_2^1  = N_1^1  - L_1^1  + 1
\end{equation} 

\begin{figure*}[!t]
\centering
\includegraphics[scale=1]{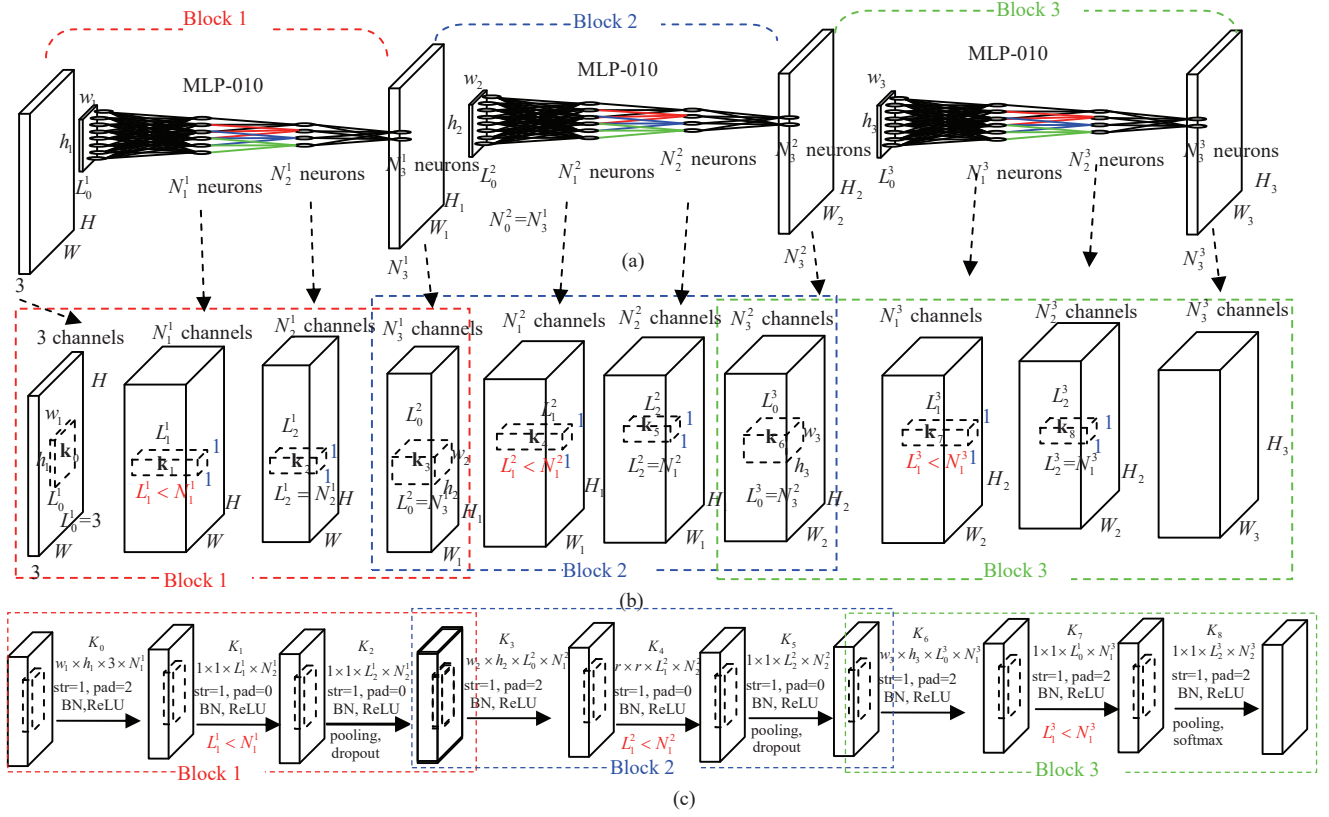}
\caption{ The architecture of CiC-1D. (a) Directly showing the role of MLP-010. (b) The kernels and their constraints for implementing CiC-1D. (c) The architecture and main steps of CiC-1D.}
\label{Fig2}
\end{figure*}

\begin{figure*}[!t]
\centering
\includegraphics[scale=1]{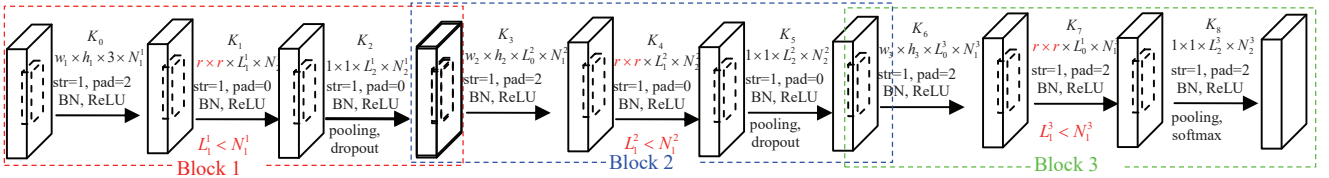}
\caption{The architecture of CiC-3D.}
\label{Fig3}
\end{figure*}
 
The success of CiC is own to the following two factors related to the second kernel in each block: (1) The convolutional manner of sparsely connecting different channels in channel dimension; and (2) The sparse connection is accomplished by unshared convolution. 

\textbf{The third kernel} ${{\bf{k}}_2}$(${\bf{K}}_2$): ${\bf{k}}_2 \in R^{1 \times 1 \times L_2^1 }$ (or ${\bf{K}}_2  \in R^{1  \times 1  \times L_2^1  \times N_3^1 }$) is also a ${1 \times 1}$ filter in spatial domain. The length ${L_2^1}$ of the filer in channel dimension is equal to the number ${N_2^1}$ of the input channels (i.e., ${L_2^1 = N_2^1}$), meaning that the channels corresponding to the same spatial location are fully connected. 

The output of the block 1 is used as the input of the block 2. The computation process of block 2 and block 3 is similar to that of block 1. The sizes and constraints of the three and four order tensors are given in Table I. 

\begin{table*}[!t]
\renewcommand{\arraystretch}{1.5}
\caption{ The sizes and constraints of the kernels (is a three-order tensor and ${\bf{K}}_i$ is its corresponding four-order tensor)}
\label{Tab1}
\centering
\begin{tabular}{|c|c|c|c|c|c|c|}
\hline
\multirow{3}*{\rotatebox{-90}{block 1}} 
& ${{\bf{k}}_0}$ & ${\bf{K}}_0$ & ${{\bf{k}}_1}$ & ${\bf{K}}_1$ & ${{\bf{k}}_2}$ & ${\bf{K}}_2$ \\
\cline{2-7} 
& ${w_1 \!\times h_1 \!\times L_0^1}$ & ${w_1 \!\times h_1 \!\times L_0^1 \!\times N_1^1}$ & ${1 \!\times 1 \!\times L_1^1}$ & ${1 \!\times 1 \!\times L_1^1 \!\times N_2^1}$ & ${1 \!\times 1 \!\times L_2^1}$ & ${1 \!\times 1 \!\times L_2^1 \!\times N_3^1}$ \\
\cline{2-7} 
& \multicolumn{2}{|c|}{${L_0^1 = 3}$} & \multicolumn{2}{|c|}{${L_1^1 < N_1^1}$} & \multicolumn{2}{|c|}{${L_2^1 = N_2^1}$} \\

\hline
\multirow{3}*{\rotatebox{-90}{block 2}} 
& ${{\bf{k}}_3}$ & ${\bf{K}}_3$ & ${{\bf{k}}_4}$ & ${\bf{K}}_4$ & ${{\bf{k}}_5}$ & ${\bf{K}}_5$ \\
\cline{2-7} 
& 1 & 2 & 3 & 4 & 5 & 6 \\
\cline{2-7} 
& \multicolumn{2}{|c|}{${L_0^2 = N_3^1}$} & \multicolumn{2}{|c|}{${L_1^2 < N_1^2}$} & \multicolumn{2}{|c|}{${L_2^2 < N_2^2}$} \\

\hline
\multirow{3}*{\rotatebox{-90}{block 3}} 
& ${{\bf{k}}_6}$ & ${\bf{K}}_6$ & ${{\bf{k}}_7}$ & ${\bf{K}}_7$ & ${{\bf{k}}_8}$ & ${\bf{K}}_8$ \\
\cline{2-7} 
& 1 & 2 & 3 & 4 & 5 & 6 \\
\cline{2-7} 
& \multicolumn{2}{|c|}{${L_0^3 = N_3^2}$} & \multicolumn{2}{|c|}{${L_1^3 < N_1^3}$} & \multicolumn{2}{|c|}{${L_2^3 < N_2^3}$} \\
\hline
\end{tabular}
\end{table*}

Batch Normalization (BN) \cite{Ioffe_Batchnormalization_CoRR2015} is adopted to normalize the convolutional layers. ReLU (Rectified Linear Unit) nonlinearilty \cite{Krizhevsky_Imagenetclassification_NIPS2012,Nair_Rectifiedlinear_NIPS2012} is used for model a neuron's output. Max-pooling is applied on the results of ReLU. Moreover, dropout \cite{Krizhevsky_Imagenetclassification_NIPS2012,Hinton_Improvingneural_CoRR2012} is also conducted. Fig. 2(c) shows the main steps of the proposed CiC method where ``str" and ``pad" stand for the convolution stride and padding pixels. 

\subsection{Generalized CiC with Three-dimensional Filtering across Channel-Spatial Domain: CiC-3D}
It can be seen from Section III.B that the second kernel (i.e., ${\bf{K}}_1  \in R^{1  \times 1  \times L_1^1  \times N_2^1 }$,  ${\bf{K}}_4  \in R^{1  \times 1  \times L_1^2  \times N_2^2 }$, and  ${\bf{K}}_7  \in R^{1  \times 1  \times L_1^3  \times N_2^3 }$) in each block is the key of the proposed CiC where MLP-010 is adopted. These kernels perform ${1  \times 1}$ convolutions in spatial domain and one-dimensional convolutions in the channel dimension. Hence, the convolutions in different spatial locations are independently implemented. In this Section, we propose to breakthrough this independence by changing the ${1 \times 1}$ convolutions to ${n \times n}$ convolutions with ${n > 1}$. Accordingly, in the generalized CiC (called CiC-3D), the sizes of ${{\bf{K}}_1}$, ${{\bf{K}}_4}$, and ${{\bf{K}}_7}$ are changed from ${1 \times 1 \times L_1^1 \times N_2^1}$, ${1 \times 1 \times L_1^2 \times N_2^2}$, and ${1 \times 1 \times L_1^3 \times N_2^3}$ to  ${n \times n \times L_1^1 \times N_2^1}$, ${n \times n \times L_1^2 \times N_2^2}$, and ${n \times n \times L_1^3 \times N_2^3}$, respectively. Fig. 3 shows the architecture of the proposed CiC-3D. Generally, ${n = 5}$ is able to yield good performance.

\section{Experimental Results}

We call the method proposed in Section III.B CiC-1D and the method proposed in Section III.C CiC-3D. In this section, we compare the proposed method with NiN \cite{Lin_NiN_CoRR2013}, Maxout Network \cite{Goodfellow_Maxoutnetworks_CoRR2013}, Probabilistic Maxout Network \cite{Springenberg_Improvingdeepneural_CoRR2013}, Deeply Supervised Network \cite{Lee_Deeplysupervised_CoRR2014}, and NiN+LA units \cite{Agostinelli_Learningactivation_CoRR2014} on the CIFAR10 dataset \cite{Krizhevsky_Learningmultiple_2009}, the CIFAR100 dataset \cite{Krizhevsky_Learningmultiple_2009}, and their augmented versions. The proposed methods are implemented using the MatConvNet toolbox \cite{Vedaldi_MatConvNet_ACMM2015}.

\subsection{Datasets}
The CIFAR10 dataset contains 10 classes of images with 6,000 images per class and 60,000 images in total. Among the 60,000 images, 50,000 ones are used for training and the rest 10,000 ones are used for testing. All the images have three-color channels and the image size is ${32 \times 32}$. 

By randomly flipping each training images, we obtain an augmented dataset which we call CIFAR10+. CIFAR10+ is used for tuning the parameters and showing the intermediate results. Moreover, we obtain a much larger augmented dataset (called CIFAR10++) by padding 4 pixels on each side and then randomly cropping and flipping on the fly. 

The CIFAR100 dataset \cite{Shao_Learningdeepand_TNN2014} has the same number of training images and testing images as the CIFAR10. The difference is that the CIFAR100 contains 100 instead of 10 classes. Therefore, the number of images in each class is only one tenth of that of CIFAR10. The 100 classes are grouped into 20 superclasses. Each image has two labels. One is the ``fine" label indicating the specific class and the other one is the ``coarse" label indicating the super-class. CIFAR100 is much challenging than CIFAR10 for classification.

\subsection{Configurations and Intermediate Results on the CIFAR10+ Dataset}

The CIFAR10+ dataset is used for parameter selection and the selected parameters are to be used for evaluating the proposed method on the CIFAR10, CIFAR10++, and CIFAR100 datasets. To quickly obtain the optimal parameters, only 60 training epochs are employed. It is note that 230 epochs are adopted in Sections IV.C, IV.D, and IV.E.

The sizes (parameters) of the kernels ${\bf{K}}_i$, ${i = 0,...,8}$ determine the performance of CiC-1D and CiC-3D. It is difficult to jointly choose the optimal parameters. We experimentally choose the optimal parameters of CiC-1D in a greedily manner where a parameter varies until the optimal value is found and the other parameters are kept unchanged. The obtained optimal parameters are shared with CiC-3D. 

To investigate the influence of the length of the kernel in the channel dimension of CiC-1D, in Table II, MLP-010 is used for constructing block 2 (where sparsity holds because of ${L_1^2 < N_1^2}$) and dense MLPs are used for constructing block 1 and block 3 (where ${L_1^1 = N_1^1}$ and ${L_2^1 = N_2^1}$ hold for block 1 and ${L_1^3 = N_1^3}$ and ${L_2^3 = N_2^3}$ hold for block 3). The kernels (${\bf{K}}_0$, ${\bf{K}}_1$, ${\bf{K}}_2$) in bock 1 and the kernels (${\bf{K}}_6$, ${\bf{K}}_7$, ${\bf{K}}_8$) in block 3 are fixed. In block 2, the size of ${\bf{K}}_4$ is expressed as ${1 \times 1 \times L_1^2 \times N_2^2}$ where ${L_1^2}$ is the kernel length in the channel dimension. Several values of ${L_1^2}$ are used with ${N_2^2}$ and ${L_1^3}$ changing with ${L_1^2}$ according to the ${N_2^2  = N_2^1  - L_1^2  + 1}$ and ${L_1^2 = N_2^2}$, respectively (i.e., valid convolution without zero-padding). The corresponding test error rates are given in the bottom of Table II and are visualized in Fig. 4. The red numbers in Table II are the various ${L_1^2}$ and the blue numbers are the corresponding ${N_2^2}$ and ${N_1^2}$.  

\begin{table*}[!t]
\renewcommand{\arraystretch}{1.5}
\caption{ Test error rate (\%) with different kernel length ${L_1^2}$ in channel dimension of ${\bf{K}}_4$ in block 2 of CiC-1D. Note that MLP-010 is applied in only block 2 and dense MLP is applied in both block 1 and block 3. }
\label{Tab2}
\centering
\begin{tabular}{|c|c|c|c|c|c|c|}
\hline
\multirow{1}*{} 
&  & ${L_1^2 = 3}$ & ${L_1^2 = 6}$ & ${L_1^2 = 12}$ & ${L_1^2 = 24}$ & ${L_1^2 = 48}$ \\
\hline
\multirow{3}*{\rotatebox{-90}{block 1}} 
& ${\bf{K}}_0$ & ${5 \!\times 5 \!\times3 \!\times 192}$ & ${5 \!\times 5 \!\times3 \!\times 192}$ & ${5 \!\times 5 \!\times3 \!\times 192}$ & ${5 \!\times 5 \!\times3 \!\times 192}$ & ${5 \!\times 5 \!\times3 \!\times 192}$ \\
\cline{2-7} 
& ${\bf{K}}_1$ & ${1 \!\times 1 \!\times 192 \!\times 192}$ & ${1 \!\times 1 \!\times 192 \!\times 192}$ & ${1 \!\times 1 \!\times 192 \!\times 192}$ & ${1 \!\times 1 \!\times 192 \!\times 192}$ & ${1 \!\times 1 \!\times 192 \!\times 192}$ \\
\cline{2-7} 
& ${\bf{K}}_2$ & ${1 \!\times 1 \!\times 192 \!\times 192}$ & ${1 \!\times 1 \!\times 192 \!\times 192}$ & ${1 \!\times 1 \!\times 192 \!\times 192}$ & ${1 \!\times 1 \!\times 192 \!\times 192}$ & ${1 \!\times 1 \!\times 192 \!\times 192}$\\
\hline
\multirow{3}*{\rotatebox{-90}{block 2}} 
& ${\bf{K}}_3$ & ${5 \!\times 5 \!\times 192 \!\times 192}$ & ${5 \!\times 5 \!\times 192 \!\times 192}$ & ${5 \!\times 5 \!\times 192 \!\times 192}$ & ${5 \!\times 5 \!\times 192 \!\times 192}$ & ${5 \!\times 5 \!\times 192 \!\times 192}$ \\
\cline{2-7} 
& ${\bf{K}}_4$ & ${1 \!\times 1 \!\times \color{red}{3} \!\times \color{blue}190}$ & ${1 \!\times 1 \!\times \color{red}6 \!\times \color{blue}187}$ & ${1 \!\times 1 \!\times \color{red}12 \!\times \color{blue}181}$ & ${1 \!\times 1 \!\times \color{red}24 \!\times \color{blue}169}$ & ${1 \!\times 1 \!\times \color{red}48 \!\times \color{blue}143}$ \\
\cline{2-7} 
& ${\bf{K}}_5$ & ${1 \!\times 1 \!\times {\color{blue}190} \!\times 192}$ & ${1 \!\times 1 \!\times {\color{blue}187} \!\times 192}$ & ${1 \!\times 1 \!\times {\color{blue}181} \!\times 192}$ & ${1 \!\times 1 \!\times {\color{blue}169} \!\times 192}$ & ${1 \!\times 1 \!\times {\color{blue}143} \!\times 192}$ \\
\hline
\multirow{3}*{\rotatebox{-90}{block 3}} 
& ${\bf{K}}_6$ & ${3 \!\times 3 \!\times 192 \!\times 192}$ & ${3 \!\times 3 \!\times 192 \!\times 192}$ & ${3 \!\times 3 \!\times 192 \!\times 192}$ & ${3 \!\times 3 \!\times 192 \!\times 192}$ & ${3 \!\times 3 \!\times 192 \!\times 192}$ \\
\cline{2-7} 
& ${\bf{K}}_7$ & ${1 \!\times 1 \!\times 192 \!\times 192}$ & ${1 \!\times 1 \!\times 192 \!\times 192}$ & ${1 \!\times 1 \!\times 192 \!\times 192}$ & ${1 \!\times 1 \!\times 192 \!\times 192}$ & ${1 \!\times 1 \!\times 192 \!\times 192}$ \\
\cline{2-7} 
& ${\bf{K}}_8$ & ${1 \!\times 1 \!\times 192 \!\times 10}$ & ${1 \!\times 1 \!\times 192 \!\times 10}$ & ${1 \!\times 1 \!\times 192 \!\times 10}$ & ${1 \!\times 1 \!\times 192 \!\times 10}$ & ${1 \!\times 1 \!\times 192 \!\times 10}$ \\
\hline
\multicolumn{2}{|c|}{error rate (\%)} & 9.07 & 9.15 & 9.15 & 9.47 & 9.68 \\
\hline
\end{tabular}

\end{table*}

From Fig. 4, one can find that small kernel length ${L_1^2 = 3}$ in channel dimension results in the lowest error rate 9.07\%. \textbf{So we choose 3 as the optimal kernel length in channel dimension whenever MLP-010 is used for sparse connection.}  

\begin{figure}[!t]
\centering
\includegraphics[scale=0.6]{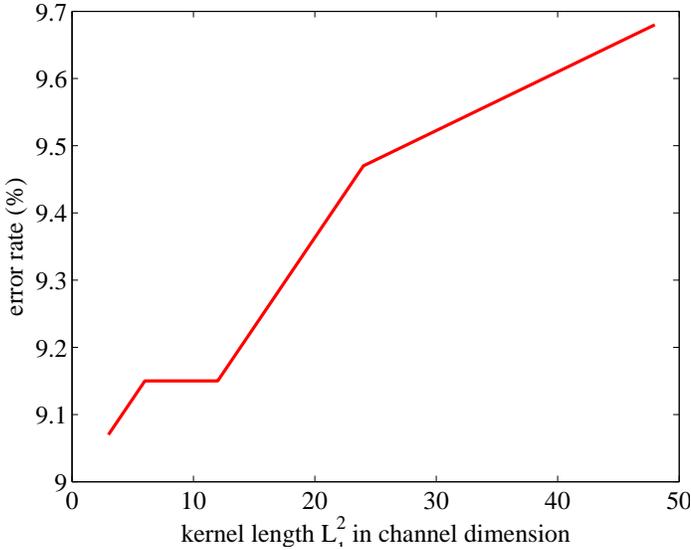}
\caption{Test error rates (\%) with different kernel length ${L_1^2}$ in channel dimension of ${\bf{K}}_4$ in block 2 of CiC-1D. }
\label{Fig4}
\end{figure}

The number ${N_1^2}$ of the input channels of ${\bf{K}}_4$ in block 2 is also important for classification. To seek the optimal value of ${N_1^2}$, we fix the kernel length ${L_1^2 = 3}$ in channel dimension and utilize different ${N_1^2}$ (i.e., 160, 192, 224, and 256). The resulting classification error rates on the CIFAR10+ dataset are given in Table III and shown in Fig. 5. \textbf{It is observed that the best classification performance is obtained when the number of input channels of an MLP-010 is 224.} 

\begin{table*}[!t]
\renewcommand{\arraystretch}{1.5}
\caption{ Test error rate (\%) vary with the number ${N_1^2}$ (marked in red color) of the input channels of ${\bf{K}}_4$ in block 2 when the kernel length ${L_1^2}$ in channel dimension of ${\bf{K}}_4$ is fixed ${L_1^2 = 3}$ (marked in green color).}
\label{Tab3}
\centering
\begin{tabular}{|c|c|c|c|c|c|}
\hline
\multirow{1}*{} 
&  & ${N_1^2 = 160}$ & ${N_1^2 = 192}$ & ${N_1^2 = 224}$ & ${N_1^2 = 256}$  \\
\hline
\multirow{3}*{\rotatebox{-90}{block 1}} 
& ${\bf{K}}_0$ & ${5 \!\times 5 \!\times3 \!\times 192}$ & ${5 \!\times 5 \!\times3 \!\times 192}$ & ${5 \!\times 5 \!\times3 \!\times 192}$ & ${5 \!\times 5 \!\times3 \!\times 192}$ \\
\cline{2-6} 
& ${\bf{K}}_1$ & ${1 \!\times 1 \!\times 192 \!\times 192}$ & ${1 \!\times 1 \!\times 192 \!\times 192}$ & ${1 \!\times 1 \!\times 192 \!\times 192}$ & ${1 \!\times 1 \!\times 192 \!\times 192}$  \\
\cline{2-6} 
& ${\bf{K}}_2$ & ${1 \!\times 1 \!\times 192 \!\times 192}$ & ${1 \!\times 1 \!\times 192 \!\times 192}$ & ${1 \!\times 1 \!\times 192 \!\times 192}$ & ${1 \!\times 1 \!\times 192 \!\times 192}$\\
\hline
\multirow{3}*{\rotatebox{-90}{block 2}} 
& ${\bf{K}}_3$ & ${5 \!\times 5 \!\times 192 \!\times \color{red}160}$ & ${5 \!\times 5 \!\times 192 \!\times \color{red}192}$ & ${5 \!\times 5 \!\times 192 \!\times \color{red}224}$ & ${5 \!\times 5 \!\times 192 \!\times \color{red}256}$ \\
\cline{2-6} 
& ${\bf{K}}_4$ & ${1 \!\times 1 \!\times {\color{green}{3}} \!\times 158}$ & ${1 \!\times 1 \!\times {\color{green}{3}} \!\times 190}$ & ${1 \!\times 1 \!\times {\color{green}{3}} \!\times 222}$ & ${1 \!\times 1 \!\times {\color{green}{3}} \!\times 254}$  \\
\cline{2-6} 
& ${\bf{K}}_5$ & ${1 \!\times 1 \!\times 158 \!\times 192}$ & ${1 \!\times 1 \!\times 190 \!\times 192}$ & ${1 \!\times 1 \!\times 222 \!\times 192}$ & ${1 \!\times 1 \!\times 254 \!\times 192}$ \\
\hline
\multirow{3}*{\rotatebox{-90}{block 3}} 
& ${\bf{K}}_6$ & ${3 \!\times 3 \!\times 192 \!\times 192}$ & ${3 \!\times 3 \!\times 192 \!\times 192}$ & ${3 \!\times 3 \!\times 192 \!\times 192}$ & ${3 \!\times 3 \!\times 192 \!\times 192}$ \\
\cline{2-6} 
& ${\bf{K}}_7$ & ${1 \!\times 1 \!\times 192 \!\times 192}$ & ${1 \!\times 1 \!\times 192 \!\times 192}$ & ${1 \!\times 1 \!\times 192 \!\times 192}$ & ${1 \!\times 1 \!\times 192 \!\times 192}$ \\
\cline{2-6} 
& ${\bf{K}}_8$ & ${1 \!\times 1 \!\times 192 \!\times 10}$ & ${1 \!\times 1 \!\times 192 \!\times 10}$ & ${1 \!\times 1 \!\times 192 \!\times 10}$ & ${1 \!\times 1 \!\times 192 \!\times 10}$  \\
\hline
\multicolumn{2}{|c|}{error rate (\%)} & 9.42 & 9.07 & \textbf{8.97} & 9.33  \\
\hline
\end{tabular}

\end{table*}

\begin{figure}[!t]
\centering
\includegraphics[scale=0.6]{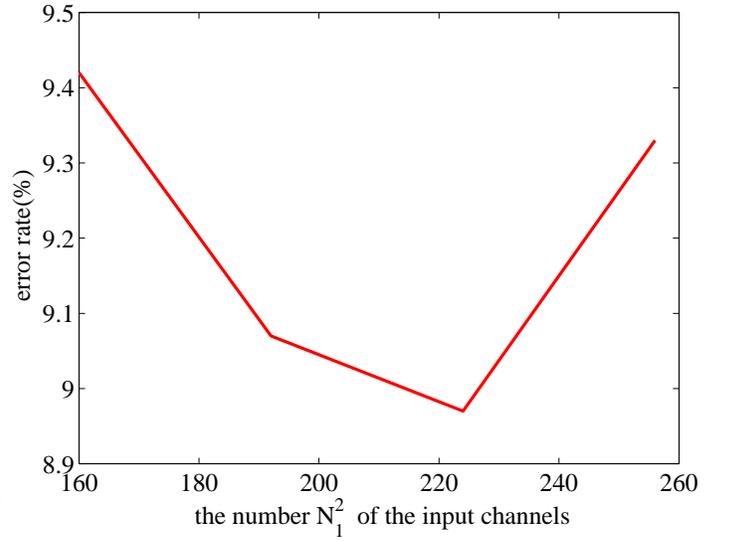}
\caption{Test error rates versus the number ${N_1^2}$ of the input channels of ${\bf{K}}_4$ in block 2. }
\label{Fig5}
\end{figure}

In Table II and Table III, MLP-010 is applied only in block 2 and dense MLPs are employed in block 1 and block 3. To further investigate the importance of MLP-010, MLP-010 is applied in both block 1 and block 2 while dense MLP is applied in block 3. Moreover, all the three blocks can employ MLP-010. Note that the number of the input channels of the sparsely connected layer of MLP-010 is 224 and the kernel length in channel dimension is 3. The corresponding results are given in the second row of Table IV from which one can observe that it is beneficial to apply MLP-010 in more blocks. The third row of Table IV shows the test error rates of CiC-3D where the number of the input channels of the sparsely connected layer of MLP-010 is also 224 and the kernel length in channel dimension is also 3. For CiC-3D, it is also desirable to apply MLP-010 in all the three blocks.

\begin{table}[!t]
\renewcommand{\arraystretch}{1.5}
\caption{Test error rates (\%) when MLP-010 is applied in only block 1, block 1 and block 2, and all the three blocks, respectively.}
\label{Tab4}
\centering
\begin{tabular*}{8cm}{@{\extracolsep{\fill}}lcccc}
\hline
 & only block 2 & block 1 and 2 & block 1, 2, and 3\\
\hline
CiC-1D       &9.23   &---     &---\\
CiC-3D       &9.05   &8.46  &7.94 \\
\hline
\end{tabular*}
\end{table}

Comparing the second row and the third row of Table III, one can conclude that CiC-3D outperforms CiC-1D significantly. So in the following experiments, only CiC-3D is employed with MLP-010 being used in all the three blocks. The main parameters and operation of the proposed CiC-3D are shown in Fig. 6. 

In Fig. 6, ``str" and ``pad" stands for the convolution stride and padding pixels. ``BN" and ReLU mean Batch Normalization \cite{Ioffe_Batchnormalization_CoRR2015} and Rectified Linear Units \cite{Krizhevsky_Imagenetclassification_NIPS2012}, respectively. ``Pooling ${3 \times 3}$" and ``pooling ${8 \times 8}$" mean that max pooling are conducted with ${3 \times 3}$ template and ${8 \times 8}$ template, respectively.

\begin{figure}[!t]
\centering
\includegraphics[scale=1]{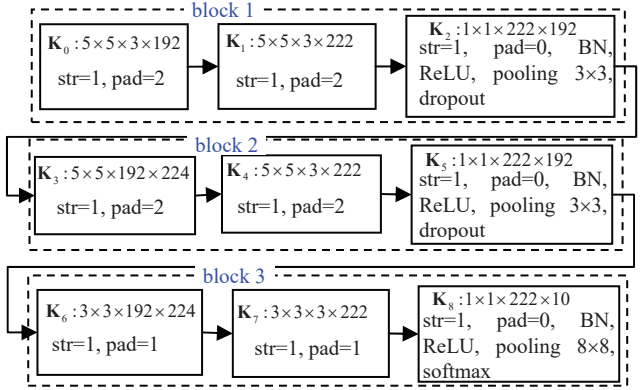}
\caption{Configuration of the proposed CiC-3D.}
\label{Fig6}
\end{figure}

The learning rates of a CNN is important for training. The learning rates in the first 80 epochs are identical to 0.5. From the 81-th epochs to the 180-th epochs, the learning rate decreases from 0.5 to 0.005 with step -0.005. From the 181-th epochs to the 230-th epochs, the learning rate decreases from 0.005 to 0.00005 with step -0.0001. Table V shows the learning rates of different training epochs. 

\begin{table}[!t]
\renewcommand{\arraystretch}{1.5}
\caption{Learning rates used for training CiC}
\label{Tab5}
\centering
\begin{tabular*}{8cm}{@{\extracolsep{\fill}}lcccc}
\hline
Epoch interval & [1, 80] & [81, 180] & [181, 230]\\
\hline
From       &0.5   &0.5     &0.005\\
To         &0.5   &0.005   &0.0005 \\
Step       &0     &-0.005  &-0.0001\\
\hline
\end{tabular*}
\end{table}

\subsection{Comparison with Other Methods on the CIFAR10 Dataset}
In Section IV.B, only 60 training epochs are employed. Hereinafter, the training epochs of CiC-3D is up to 230 epochs. Note that the CIFAR10+ dataset is used in Section IV.B and the original CIFAR10 dataset is used in this section.

We first show in Fig. 7 the curves of training error rates vs. training epochs of NiN \cite{Lin_NiN_CoRR2013} (see Fig. 8 for its architecture) and the proposed CiC-3D (see Fig. 6) on the CIFAR-10 dataset. It is observed that CiC-3D has smaller training error rates than NiN. Moreover, the proposed method CiC-3D convergences much faster than NiN. 

\begin{figure}[!t]
\centering
\includegraphics[scale=0.5]{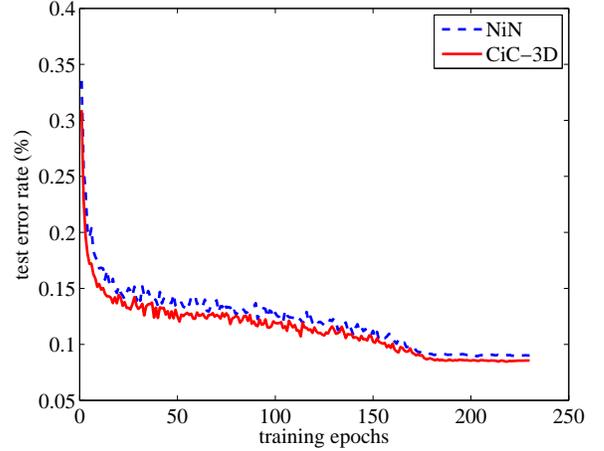}
\caption{Training error (\%) vs. iteration of NiN and CiC-3D on the CIFAR-10 dataset.}
\label{Fig7}
\end{figure}

\begin{figure}[!t]
\centering
\includegraphics[scale=1]{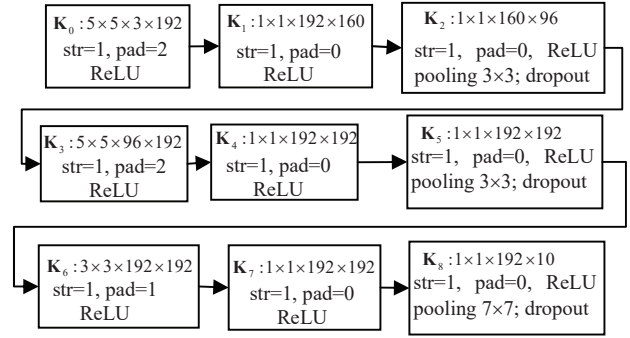}
\caption{Configuration of NiN \cite{Lin_NiN_CoRR2013} for the CIFAR10 dataset.}
\label{Fig8}
\end{figure}

Table VI gives the test error rates of NiN \cite{Lin_NiN_CoRR2013}, Deeply Supervised Network (DSN) \cite{Lee_Deeplysupervised_CoRR2014}, NiN-LA units (NiN-LA)\cite{Agostinelli_Learningactivation_CoRR2014}, RCNN-160\cite{Liang_Recurrent_CVPR2015}, and CiC-3D on the CIFAR10 dataset. One can see from Table VI that CiC-3D outperforms NiN by 1.95 percent. In addition, CiC-3D gives 1.13 percent improvement over the NiN-LA. 

\begin{table}[!t]
\renewcommand{\arraystretch}{1.5}
\caption{Test error rates (\%) on the CIFAR10 dataset}
\label{Tab6}
\centering
\begin{tabular*}{8.5cm}{@{\extracolsep{\fill}}lccccc}
\hline
 NiN\cite{Lin_NiN_CoRR2013}& DSN\cite{Lee_Deeplysupervised_CoRR2014} & NiN-LA\cite{Agostinelli_Learningactivation_CoRR2014} & RCNN-160\cite{Liang_Recurrent_CVPR2015} & CiC-3D\\
\hline
10.41 &9.78 &9.59 & 8.69 &\textbf{8.46}\\
\hline
\end{tabular*}
\end{table}

\subsection{Comparison with Other Methods on the CIFAR10++ Dataset}
As stated in Section IV.A, the CIFAR10++ dataset is a large version of CIFAR10. The configuration of CiC-3D is the same as that in Section IV.C (i.e., Fig. 6). 

\begin{table}[!t]
\renewcommand{\arraystretch}{1.5}
\caption{Test error rates (\%) on the CIFAR10${++}$ dataset}
\label{Tab7}
\centering
\begin{tabular*}{8.5cm}{@{\extracolsep{\fill}}lccccc}
\hline
 NiN\cite{Lin_NiN_CoRR2013}& ResNet-1202\cite{He_Deepresidual_CoRR2015} &NiN-LA\cite{Agostinelli_Learningactivation_CoRR2014} &  RCNN-160\cite{Liang_Recurrent_CVPR2015} & CiC-3D\\
\hline
8.81 &7.93 &7.51  & 7.09 &\textbf{6.68}\\
\hline
\end{tabular*}
\end{table}

The test error rates on the CIFAR10++ dataset is given in Table VII. The test error rates of NiN and CiC-3D are 8.81\% and 6.68\%, respectively. So CiC-3D gives 2.13\% percent improvement over NiN. The superiority of CiC-3D grows as the training set increases. 

\subsection{Comparison with Other Methods on the CIFAR100 Dataset}
The CIFAR100 dataset is challenging than the CIFAR10 dataset. But we still adopt the configuration in Fig. 6 for constructing CiC-3D. 

\begin{table}[!t]
\renewcommand{\arraystretch}{1.5}
\caption{Test error rates (\%) on the CIFAR100 dataset without augment}
\label{Tab7}
\centering
\begin{tabular*}{8.5cm}{@{\extracolsep{\fill}}lcccccc}
\hline
 NiN\cite{Lin_NiN_CoRR2013}&  NiN-LA\cite{Agostinelli_Learningactivation_CoRR2014} & Highway\cite{Srivastava_Highwaynetworks_CoRR2015} & RCNN-160\cite{Liang_Recurrent_CVPR2015} & CiC-3D\\
\hline
35.68 &34.40 & 32.24 & 31.75 &\textbf{31.40}\\
\hline
\end{tabular*}
\end{table}

Table VIII shows that the test error rates of NiN \cite{Lin_NiN_CoRR2013}, NiN+LA units (NiN-LA)\cite{Agostinelli_Learningactivation_CoRR2014}, Highway\cite{Srivastava_Highwaynetworks_CoRR2015}, RCNN-160\cite{Liang_Recurrent_CVPR2015} and CiC-3D are 35.68\%, 34.40\%, 32.24\%, 31.75\%, and 31.40\%, respectively. CiC-3D arrives at the lowest test error rate and outperforms NiN and NiN-LA by 4.28 percent and 3 percent, respectively. 

The above experimental results show that the proposed CiC-3D method significantly outperforms NiN in reducing the test error rate.

\section{Conclusion and Future Work}
In this paper, we have presented a CNN method (called CiC) where sparse shallow MLP is used for convolution. Full sparse MLP and several types of partial sparse MLPs (e.g., MLP-010, MLP 011, MLP-100, MLP-110) were proposed. The MLP-010 was employed in the experiments. The main idea is to sparsely connect different channels in a unshared convolutional manner. The basic version of CiC is CiC-1D and its generalized version is CiC-3D. In CiC-1D, a one-dimensional filter is employed for connecting the second layer of each block. CiC-1D was then generalized to CiC-3D by utilizing a three dimensional filter across channel-spatial domain.

In the experiments, a partial spare MLP, MLP-010, was adopted. In the future, full sparse MLP and other types of sparse MLPs can be implemented. Moreover, the proposed idea can be integrated to other state-of-the-art CNNs so that better results can be expected. 

\ifCLASSOPTIONcaptionsoff
  \newpage
\fi


\begin{thebibliography}{1}
%
  
\bibitem{Bengio_Learningdeeparchitectures_FTML2006}  
Y. Bengio, ``Learning deep architectures for AI,"
\textit{Foundations and Trends in Machine Learning}, vol. 2, no. 1, pp. 1-127, 2006. 

\bibitem{Goodfellow_DeepLearning_2015}  
I. Goodfellow, A. Courville, and Y. Bengio, Deep Learning,
\textit2015. 

\bibitem{Lin_NiN_CoRR2013}  
M. Lin, Q. Chen, and S. Yan, ``Network in network,"
\textit{CoRR}, abs/1312.4400, 2013. 

\bibitem{Haykin_NeuralNetworks_1999}  
S. Haykin, Neural Networks: A Comprehensive Foundation,
\textit Second Edition, Prentic Hall,1999. 

\bibitem{LeCun_Gradientbased_IEEE1998}  
Y. LeCun, L. Bottou, Y. Bengio, and P. Haffner, ``Gradient-based learning applied to document recognition,"
\textit{Proceedings of the IEEE}, vol. 86, no. 11, pp. 2278-2324, 1998. 

\bibitem{Krizhevsky_Imagenetclassification_NIPS2012}  
A. Krizhevsky, I. Sutskever, and G. Hinton, ``Imagenet classification with deep convolutional neural networks," in  
\textit{Proc. Advances in Neural Information Processing Systems}, 2012, pp. 1097-1105. 

\bibitem{Szegedy_Goingdeeper_CoRR2014}  
C. Szegedy, W. Liu, Y. Jia, P. Sermanet, S. Reed, D. Anguelov, D. Erhan, V. Vanhoucke, and A. Rabinovich, ``Going deeper with convolutions," 
\textit{CoRR}, abs/1409.4842, 2014. 

\bibitem{Simonyan_Verydeep_CoRR2014}  
K. Simonyan and A. Zisserman, ``Very deep convolutional networks for large-scale image recognition," 
\textit{CoRR}, abs/1409.1556, 2014. 

\bibitem{He_Deepresidual_CoRR2015}  
K. He, X. Zhang, S. Ren, and J. Sun, ``Deep residual learning for image recognition," 
\textit{CoRR}, abs/1512.03385, 2015.    

\bibitem{Srivastava_Highwaynetworks_CoRR2015}  
R. K. Srivastava, K. Greff, and J. Schmidhuber, ``Highway networks," 
\textit{CoRR}, abs/1505.00387, 2015.  

\bibitem{Sermanet_Overfeat_CoRR2013}  
P. Sermanet, D. Eigen, X. Zhang, M. Mathieu, R. Fergus, and Y. LeCun, ``Overfeat: integrated recognition, localization and detection using convolutional networks," 
\textit{CoRR}, abs/1312.6229, 2013.

\bibitem{Matthew_Visualizingand_ECCV2014}  
Matthew D. Zeiler and Rob Fergus, ``Visualizing and understanding convolutional networks," in
\textit{Proc. European Conf. Computer Vision}, 2014.

\bibitem{LeCun_Convolutionalnetworks_ISCS2010}  
Y. LeCun, K. Kavukcuoglu, and C. Farabet, ``Convolutional networks and applications in vision," in
\textit{Proc. IEEE International Symposium on Circuits and Systems}, 2010. 

\bibitem{Jarrett_Whatis_ICCV2009}  
K. Jarrett, K. Kavukcuoglu, M. A. Ranzato, and Y. LeCun, ``What is the best multi-stage architecture for object recognition?" in
\textit{Proc IEEE International Conference on Computer Vision}, 2009, pp. 2146-2153. 

\bibitem{Ciresan_Multicolumndeep_CoRR2012}  
D. Ciresan, U. Meier, and J. Schmidhuber, ``Multi-column deep neural networks for image classification," 
\textit{CoRR}, abs/1202.2745, 2012. 

\bibitem{Gregor_Emergenceof_CoRR2010}  
K. Gregor and Y. LeCun, ``Emergence of complex-like cells in a termporal product network with local receptive fields," 
\textit{CoRR}, abs/1006.044.8.308, 2010.

\bibitem{Zeiler_Stochasticpooling_CoRR2013}  
M.D. Zeiler and R. Fergus, ``Stochastic pooling for regularization of deep convolutional neural networks," 
\textit{CoRR}, abs/1301.3557, 2013. 

\bibitem{He_Spatialpyramid_ECCV2014}  
K. He, X. Zhang, S. Ren, and J. Sun, ``Spatial pyramid pooling in deep convolutional networks for visual recognition," in
\textit{Proc. European Conf. Computer Vision}, 2014, pp. 346-361. 

\bibitem{Chan_PCANet_CoRR2014}  
T. Chan, K. Jia, S. Gao, J. Lu, Z. Zeng, and Y. Ma, ``PCANet: a simple deep learning baseline for image classification?" 
\textit{CoRR}, abs/1404.3606, 2014.

\bibitem{Lee_Generalizingpooling_CoRR2015}  
C. Lee, P. Gallagher, and Z. Tu, ``Generalizing pooling functions in convolutional neural networks: mixed, gated, and tree," 
\textit{CoRR}, abs/1509.08985, 2015. 

\bibitem{Springenberg_Strivingfor_CoRR2014}  
J. Springenberg, A. Dosovitskiy, T. Brox T, and M. Riedmiller, ``Striving for simplicity: the all convolutional net," 
\textit{CoRR}, abs/1412.6806, 2014.

\bibitem{Gong_Multiscaleorderless_ECCV2014}  
Y. Gong, L. Wang, R. Guo, and S. Lazebnik, ``Multi-scale orderless pooling of deep convolutional activation features," in
\textit{Proc. European Conf. Computer Vision}, 2014, pp. 392-407. 

\bibitem{Yoo_Multiscalepyramid_CVPR2015}  
D. Yoo, S. Park, J. Lee, and I. Kweon, ``Multi-scale pyramid pooling for deep convolutional representation," in
\textit{Proc. IEEE Workshop Computer Vision and Pattern Recognition}, 2015.

\bibitem{Graham_FractionalMax_CoRR2014}  
B. Graham, ``Fractional Max-Pooling," 
\textit{CoRR}, abs/1412.6071, 2014.

\bibitem{Murray_Generalizedmax_ICML2010}  
N. Murray and F. Perronnin, ``Generalized max pooling," in
\textit{Proc. IEEE International Conf. Computer Vision and Pattern Recognition}, 2014, pp. 2473-2480.

\bibitem{Nair_Rectifiedlinear_NIPS2012}  
V. Nair and G. Hinton, ``Rectified linear units improve restricted Boltzmann machines," in
\textit{Proc. International Conf. Machine Learning}, 2010. 

\bibitem{Hinton_Improvingneural_CoRR2012}  
G. Hinton, N. Srivastava, A. Krizhevsky, I. Sutskever, and R.R. Salakhutdinov, ``Improving neural networks by preventing co-adaptation of feature detectors,"
\textit{ CoRR}, abs/ 1207.0580, 2012.

\bibitem{Krizhevsky_Learningmultiple_2009}  
A. Krizhevsky and G. Hinton, Learning multiple layers of features from tiny images, 
\textit Master’s thesis, Department of Computer Science, University of Toronto, 2009.

\bibitem{Goodfellow_Maxoutnetworks_CoRR2013}  
I. Goodfellow, D. Warde-Farley, M. Mirza, A. Courville, and Y. Bengio, ``Maxout networks," 
\textit{CoRR}, abs/1302.4389, 2013.

\bibitem{Springenberg_Improvingdeepneural_CoRR2013}  
JT. Springenberg, M. Riedmiller, ``Improving deep neural networks with probabilistic maxout units," 
\textit{CoRR}, abs/1312.6116, 2013. 

\bibitem{Lee_Deeplysupervised_CoRR2014}  
C. Lee, S. Xie, P. Gallagher, Z. Zhang, and Z. Tu, ``Deeply-supervised nets," 
\textit{CoRR}, abs/1409.5185, 2014. 

\bibitem{Agostinelli_Learningactivation_CoRR2014}  
F. Agostinelli, M. Hoffman, P. Sadowski, and P. Baldi, ``Learning activation functions to improve deep neural networks," 
\textit{CoRR}, abs/1412.6830, 2014.

\bibitem{Ioffe_Batchnormalization_CoRR2015}  
S. Ioffe and C. Szegedy, ``Batch normalization: accelerating deep network training by reducing internal covariate shift," 
\textit{CoRR}, abs/1502.03167, 2015.

\bibitem{Jiang_CascadedSubpatch_CoRR2016}  
X. Jiang, Y. Pang, M. Sun, and X. Li, ``Cascaded Subpatch Networks for Effective CNNs,"  
\textit{CoRR}, abs/1603.00128, 2016. 

\bibitem{Vedaldi_MatConvNet_ACMM2015}  
A. Vedaldi and K. Lenc, ``MatConvNet: convolutional neural networks for matlab," in
\textit{Proc. ACM Multimedia}, 2015, pp. 689-692.

\bibitem{Gong_AMultiobjective_TNN2015}  
Maoguo Gong, Jia Liu, Hao Li, Qing Cai, and Linzhi Su, ``A Multi objective Sparse Feature Learning Model for Deep Neural Networks," 
\textit{IEEE Transactions on Neural Networks and Learning Systems}, vol. 26, no. 12, pp. 3263-3277, 2015. 

\bibitem{Shao_Learningdeepand_TNN2014}  
Ling Shao, Di Wu, and Xuelong Li, ``Learning deep and wide: a spectral method for learning deep networks," 
\textit{IEEE Transactions on Neural Networks and Learning Systems}, vol. 25, no. 12, pp. 2303-2308, 2014.

\bibitem{Chang_Deepandshallow_TNN2015}  
Chih-Hung Chang, ``Deep and shallow architecture of multilayer neural networks," 
\textit{IEEE Transactions on Neural Networks and Learning Systems}, 2015, vol. 26, no. 10, pp. 2477-2486, 2015.

\bibitem{Liang_Recurrent_CVPR2015}  
M. Liang and X. Hu, ``Recurrent convolutional neural network for object recognition," in
\textit{Proc. IEEE Conference on Computer Vision and Pattern Recognition}, 2015, pp. 3367-3375. 


\end{thebibliography}
\end{document}